\documentclass[conference]{IEEEtran}
\IEEEoverridecommandlockouts
\usepackage{cite}
\usepackage{amsmath,amssymb,amsfonts}
\usepackage{algorithmic}
\usepackage{graphicx}
\usepackage{textcomp}
\usepackage{xcolor}
\def\BibTeX{{\rm B\kern-.05em{\sc i\kern-.025em b}\kern-.08em
    T\kern-.1667em\lower.7ex\hbox{E}\kern-.125emX}}
\begin{document}

\title{Document Understanding for Healthcare Referrals \
}

\author{\IEEEauthorblockN{Jimit Mistry}
\IEEEauthorblockA{
\textit{Infinx Healthcare}\\
Austin, USA \\
jimit.mistry@infinx.com}
\and
\IEEEauthorblockN{Natalia M. Arzeno}
\IEEEauthorblockA{
\textit{Infinx Healthcare}\\
Austin, USA \\
natalia@infinx.com}
}

\maketitle

\begin{abstract}
Reliance on scanned documents and fax communication for healthcare referrals leads to high administrative costs and errors that may affect patient care. In this work we propose a hybrid model leveraging LayoutLMv3 along with domain-specific rules to identify key patient, physician, and exam-related entities in faxed referral documents. We explore some of the challenges in applying a document understanding model to referrals, which have formats varying by medical practice, and evaluate model performance using MUC-5 metrics to obtain appropriate metrics for the practical use case. Our analysis shows the addition of domain-specific rules to the transformer model yields greatly increased precision and F1 scores, suggesting a hybrid model trained on a curated dataset can increase efficiency in referral management. 
\end{abstract}


\section{Introduction}
Despite the abundant use of electronic health record (EHR) systems, many of these systems do not communicate with each other, resulting in a healthcare industry that remains heavily reliant on fax messages \cite{Brown2021-mj}. When physicians require another provider to perform additional evaluations or procedures, a referral document is generated, and is often sent via fax or as a scanned image pdf. The receiving provider needs to understand if it pertains to an existing or new patient, who sent it, and for what reason. Specialists and imaging centers may receive hundreds of referrals a day. Faxes and scanned documents are often manually processed and entered into the corresponding patient’s medical record, creating high administrative costs and potential errors that can affect patient care. The documents sent are of varied types and qualities, from faxes corrupted with visual artifacts to generated pdfs sent as scanned files that are not parsable. In this work, we present a hybrid model for the extraction of key patient, referring physician, and exam information from referrals, and explore some of the challenges associated with this task.

In information extraction from referrals, rules-based approaches face many limitations, since each healthcare practice may have their own format for submitting referrals. Patient and physician information may or may not have corresponding \textit{Patient} and \textit{Physician} keywords near them, creating a point of failure for systems that extract key-value pairs from the text where entity classification is dependant on keys. Additionally, information about the procedure or diagnosis may be in free text form. However, referral documents often follow a general structure with patient details concentrated in one section and physician details concentrated in a different section of the page. This presents the opportunity to use models that account for both the structure of a document and the text within it. 

In this paper, we leverage LayoutLMv3 \cite{Huang2022-ch}, a multi-modal transformer model that incorporates embeddings for both image and text, along with domain-specific rules to extract entities associated with the patient, referring physician, and ordered exam from faxed or scanned referral documents. To our knowledge, there are no peer-reviewed publications for LayoutLM applications in healthcare, as a PubMed search for \textit{LayoutLM} did not retrieve any records.

\section{Related Work} 
In 2019, LayoutLM \cite{Xu2019-nx}, the first model published considering document structure as well as the text within it, successfully classified images, identified information in forms, and extracted entities from receipts. Other multi-modal models considering the image of the document as well as the text with varying architectures and pre-training objectives have followed, such as UDoc \cite{Gu2022-he}, StructuralLM \cite{Li2021-vo}, DocFormer \cite{Appalaraju2021-wn}, and GraphDoc \cite{Zhang2022-oy}, some of which outperformed LayoutLM. More recently, LayoutLMv3 was proposed \cite{Huang2022-ch} which offers state-of-the-art general performance in document understanding tasks due to its pre-training on unified text and image masking along with a word-patch alignment objective to learn crossmodal alignment. None of the datasets for these models included healthcare documents.

\section{Methods}
\subsection{Data and Annotations}
\begin{figure}[htbp]
\centerline{\includegraphics[width=0.45\textwidth]{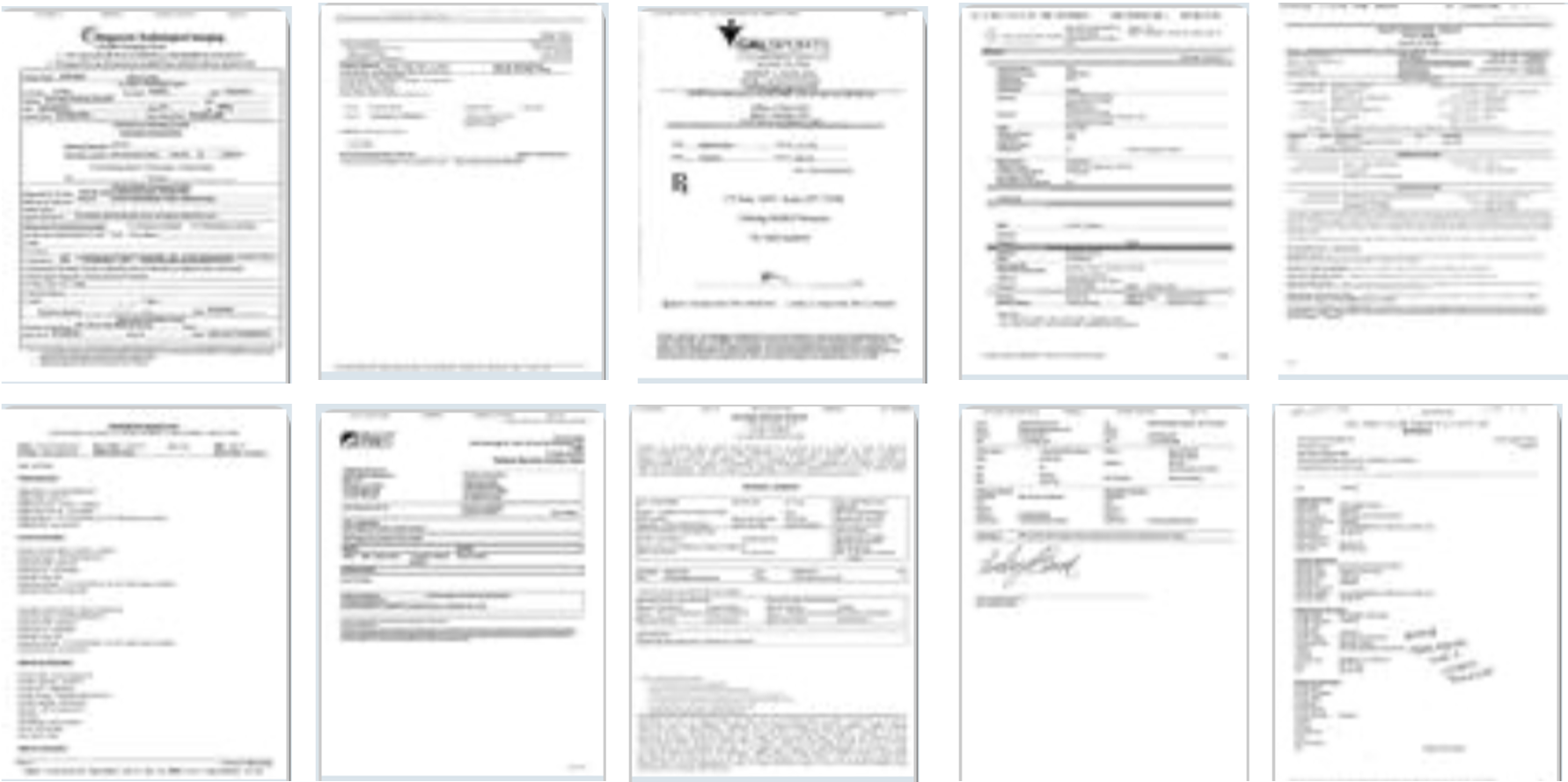}}
\caption{Sample of document structures in referrals.}
\label{fig_formats}
\end{figure}

The dataset consists of varied types and quality of scans and faxes of referrals or order documents submitted to radiology imaging centers between August and December 2022. These include faxes, scanned generated pdfs, forms with handwriting, and scans of prints, with the document structure varying based on the referring provider. Some of these documents cannot be parsed through OCR tools, and are not considered in this paper. Fig. \ref{fig_formats} illustrates some referral structures. A referral document typically consists of 1-3 pages. Model development and analyses were performed per page. 

The training set was annotated by internal team members whose day-to-day tasks include receiving scanned documents, identifying which patient they belong to in the imaging center’s system, and correctly adding them to the patient’s EHR. For the training data, the annotations were limited to a maximum of one annotation of each entity type per page. The test dataset was annotated by a separate internal team following the same guidelines with the exception that all occurrences of an entity type were annotated on a page. Documents for the test set were obtained from the same pool of scanned documents as the training set, ensuring these had not yet been annotated. 

The entity types for annotation and identification were:
\begin{enumerate}
\item Patient’s full name
\item Patient’s date of birth (DOB)
\item Patient’s gender
\item Patient’s address
\item Referring physician’s full name
\item Referring physician’s address
\item Ordered exam (procedure)
\item Reason for exam
\end{enumerate}

The training data consisted of 3032 pages and the test data consisted of 137 pages from 100 documents.

\subsection{Annotation Quality}
Fifty random samples from the train and test set were audited by both authors. Only Patient and Physician related entities were audited, as Exam-type entities can span multiple sentences and were not annotated in a consistent manner. For example, some documents have individual sentences annotated whereas others have entire sections annotated as one Exam-type entity. The average F1 score for the train and test sets after data preprocessing (Section \ref{sec:preprocessing}) was 0.92 and 0.93 respectively. Note that for the train set audit, any one instance of each entity type is considered as correct. Following this analysis, we identified 24 pages from the test data that had 3 or fewer annotations, which were removed after confirmation of incorrect or missing annotations. The final test data consisted of 112 pages.

\subsection{OCR and Text Grouping}
The images in the dataset were scans or faxes of documents, requiring OCR in order to extract the text and bounding boxes for input to LayoutLMv3. In this work, we use Amazon Textract’s \cite{Amazon_Web_Services_undated-el} text detection, which outputs text and bounding boxes at the token and line level. Given the varied structure in the referral documents, the text identified by OCR had to be additionally processed to be able to correctly order words in multi-line text. The grouping of lines is critical in order to understand entities that can span lines of text. Though we evaluated text grouping as identified by the baseline LayoutParser model \cite{Shen2021-uv}, the results were not satisfactory, likely as the structures of its training documents differ too much from the structure of referral documents.
%Unlike other OCR engines, the lines do not necessarily span the entire width of the page, but rather separate lines may be identified in the same vertical position of the page if the text is formatted in columns. 

We leverage DBSCAN \cite{Ester1996-nv}, a density-based clustering algorithm without a predetermined number of clusters, in our custom grouping of lines. Note that we follow Textract’s output of lines, which may have multiple lines for a single vertical position on the page based on the page structure.  The grouping is a multi-step process:
\begin{enumerate}
\item	Cluster lines using DBSCAN on the y-coordinate.
\item	Within each cluster, pre-define a sparse distance matrix to ensure lines are not skipped in a grouping, then run DBSCAN on the x-coordinates.
\item	Within each new cluster, span the horizontal length to detect columns and split if columns are found.
\end{enumerate}

\begin{figure}[htbp]
\centerline{\includegraphics[width=0.48\textwidth]{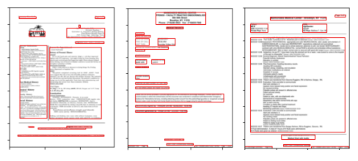}}
\caption{Grouping of OCR lines in different document structures.}
\label{fig_grouping}
\end{figure}

Fig. \ref{fig_grouping} displays sample outputs of text grouping.

\subsection{Data Preprocessing}  \label{sec:preprocessing}
Initial exploration of annotations revealed some noisy labels, thus data label corrections were implemented to improve data quality. Excess tokens in the annotations, mainly due to the field names (e.g. \emph{Name:}) were the most prevalent issue. Domain-specific stop word removal was undertaken for all entity types to remove the unwanted noise added by annotator tool limitations or annotators. Address type entities are notoriously difficult to correct because of their versatility. The stop word removal applied on address entities removed field names comprising \emph{Address}, \emph{city}, \emph{state}, \emph{line}, \emph{1:}, \emph{2:}, \emph{zip}. Additionally, phone numbers were removed from addresses and spans of some entities were corrected by searching to the left of the beginning token of the annotation, to correct single-token name annotations and addresses with street numbers or cardinal directions (N,S,E,W) missing. These and additional preprocessing correcting provider credentials and physician names incorrectly annotated as patient names (as identified via credentials) corrected 5132 labels.

\iffalse
Data Label corrections were implemented to fix the data quality issues. Firstly, domain specific stop word removal was undertaken for all entity types to remove the unwanted noise added by annotator tool limitations or annotators. Then entity type specific corrections were made. Patient Name removed all stop words that included any numbers or suffixes. Additionally, if the Patient Name had any Medical Professional related credentials immediately surrounding, or within the entity, its classification was converted to Physician Name. Next, if the name had only one token, another word was added from the left as part of the Name. Similar corrections are made for Physician Name, except when any credentials are detected surrounding the entity, it's added to be part of the Physician Name Entity. Address type entities are notoriously difficult to correct because of their versatility. The stop word removal applied on all entities removed the field names like "Address", "city', "state", "line*, "1:", "2:", "Zip", etc. On top of this, phone numbers and words related to phone numbers were removed, if they were part of the address. Sometimes, the street number and the direction letters (N,S,E,W) were missing from the start, so they were added to the address by looking to the left from the beginning of the entity. These data corrections proved fruitful as it corrected 1767\textbf{(check again)} labels.
\fi

\subsection{Model Training}\label{sec:training}
We take a pre-trained Huggingface LayoutLMv3 model \cite{huggingface_layoutlmv3} as a base model and train it on a token classification downstream task. The model expects token text, the bounding box related to the token text, and the image as the inputs. The processor for LayoutLMv3 converts them to text embeddings, 1-D positional embeddings, image section embeddings, and 2-D positional embeddings. For training, the labels have to be in the named entity recognition (NER) format consisting of B- (beginning) tokens and I- (intermediate) tokens that signify the start and continuation of an entity respectively.

%LayoutLMv3 takes bounding boxes related embeddings along with the standard text and position embeddings as input. 
The model expects a single bounding box for an entire segment of the document, with all tokens within that segment mapped to that particular bounding box. This enables the model to understand structure and context for predictions. To obtain the segments and segment-level bounding boxes, we relied on the grouping code. This has limitations, however, since grouping may result in incorrect text splits at times and hamper performance. Additionally, the entities we seek to identify often share a segment-level bounding box with other tokens and, at inference, we required the entity assignment to be at the word level given the business use case. Thus, the model had to consider word-level bounding boxes.
 
We explored three training strategies, which all used the grouping code to determine word order. The first two strategies passed only word-level bounding boxes and only segment-level bounding boxes respectively. Both models were trained for 20 epochs at a learning rate of 1-5. The third strategy split the training dataset to have 40\% data with segment-level bounding boxes and 60\% with word-level bounding boxes. The model was trained first on segment-level, and then on word-level bounding box data for 10 epochs each at 1e-5 learning rate. 

\subsection{Inference: Prediction to Complete Entity}
The inference function takes the model predictions for each token and creates a complete entity by combining contiguous B- and I- tokens of the same entity type. If a new B- token is detected, it is considered the start of a separate entity in itself. Given initial model exploration, we added the capability to consider entities starting with an I- token. Training data provided some additional challenges, as the model predicted the entities with the same issues that plagued the training data. Particularly, the disjointed entities with address labels needed to be handled in the inference function. The function first created all complete entities by combining the tokens based on the logic stated above. Then, any address entity of the same type was combined with the previously detected address entity as long as there was a margin of $\leq 5$ tokens between them. The margin of 5 tokens was calculated by identifying the maximum number of stop words between the disjointed entities (\emph{Address}, \emph{Line}, \emph{City}, \emph{State}, and \emph{Zip}, as in 
``Address line 1: \textbf{123 street number,} Address line 2: \textless empty\textgreater, City, State, Zip: \textbf{CT, ST, 12345}" where the bold words are the predicted tokens.

\subsection{Hybrid Module: Rules Functions}
 We propose a hybrid approach that applies rules to the predictions obtained from inference. Domain and entity-specific rules are applied to improve the performance of exact matches, on the entities identified by the inference function, the same logic used in preprocessing data correction functions.

\subsection{Model Evaluation}
The business case for entity extraction in referrals does not conform to traditional analyses with false positives and false negatives. For example, a patient name incorrectly identified as physician name can be considered both a false positive and a false negative for patient name, as an incorrect value is detected and the true value remains missing. The MUC-5 evaluation metrics \cite{ Chinchor1993-zo} account for such cases and are thus used for model evaluation. Each prediction (or missing prediction) is assigned to one of five classes. The classes require aligning tokens from the predictions and annotations in order to determine the degree of overlap and correctness. Let $\mathbf{\hat{t}}$ be tokens with predicted entity type $\hat{y}$, and $\mathbf{t}$ be the aligned annotated tokens with entity type $y$. The classes are:
\begin{itemize}
\item  Correct (COR): $\hat{y}=y$ AND $\mathbf{\hat{t}}=\mathbf{t}$  
\item  Partial (PAR): $\hat{y}=y$ AND $|\mathbf{\hat{t}}\cap\mathbf{t}|\geq 0.5|\mathbf{t}|$ AND $\mathbf{\hat{t}}\neq\mathbf{t}$ 
\item Incorrect (INC): $\hat{y}\neq y$  OR $|\mathbf{\hat{t}}\cap\mathbf{t}|< 0.5|\mathbf{t}|$ 
\item Missing (MIS): $y$ has no aligned predicted entity
\item Spurious (SPU): $\hat{y}$ has no aligned annotated entity
\end{itemize}

Precision and recall for the MUC-5 metrics are based on the possible and actual counts:
\begin{align*}\nonumber 
\text{possible} =& \text{ COR} + \text{PAR} + \text{INC} + \text{MIS} \\
\text{actual} =& \text{ COR} + \text{PAR} + \text{INC} + \text{SPU} 
\end{align*}
%\vspace{-18pt}
\begin{align*}
\text{partial precision} =& \: \frac{\text{COR}+0.5\times \text{PAR}}{\text{possible}} \\
\text{partial recall}=& \:\frac{\text{COR}+0.5 \times \text{PAR}}{\text{actual}}
\end{align*}

Note that the partial and incorrect predictions form part of the denominator in both the precision and recall.

Following the business use case, where the goal is to extract referral information regardless of its exact location on the page, a single prediction for each entity type is selected per page for the Patient and Physician entity types. The entity to extract is selected via majority voting with ties broken by the maximum token count in an entity. For Exam-type entities, all unique values are considered valid. Each of these selected entities is then compared against all instances of the true labels for each entity type respectively. If the selected entity matches any one instance of the same entity type from the true labels, it is considered an exact match (COR). In a partial match (PAR), the prediction should overlap with at least 50\% of the true label without an exact match. Partial precision and recall only consider 50\% of the partial matches in the numerator.

\begin{table*}[htbp]
\caption{Model Evaluation): Patient Related Entities}
\begin{center}
\begin{tabular}{|c|c|c|c|c|c|c|c|c|c|c|c|c|}
\hline
\textbf{Model}&\multicolumn{3}{|c|}{\textbf{Name}} &\multicolumn{3}{|c|}{\textbf{DOB}} &\multicolumn{3}{|c|}{\textbf{Gender}} &\multicolumn{3}{|c|}{\textbf{Address}}\\
\cline{2-13} 
\textbf{Type} & \textbf{\textit{P}}& \textbf{\textit{R}}& \textbf{\textit{F1}}& \textbf{\textit{P}}& \textbf{\textit{R}}& \textbf{\textit{F1}}& \textbf{\textit{P}}& \textbf{\textit{R}}& \textbf{\textit{F1}}& \textbf{\textit{P}}& \textbf{\textit{R}}& \textbf{\textit{F1}} \\
\hline
Base & 0.70& 0.64& 0.67& 0.92& \textbf{0.76}& \textbf{0.83}& 0.32& 0.15& 0.21& 0.53& 0.52& 0.53\\
\hline
Base+Corr& 0.71& 0.64& 0.67& 0.84& 0.59& 0.69& 0.35& \textbf{0.17}& \textbf{0.23}& 0.52& 0.55& 0.53 \\
\hline
Base+Corr+Post& \textbf{0.79}& \textbf{0.71}& \textbf{0.75}& \textbf{0.93}& 0.64& 0.76& \textbf{0.90}& 0.13& 0.22& \textbf{0.64}& \textbf{0.68}& \textbf{0.66} \\
\hline
\end{tabular}
\label{tab:patient}
\end{center}
\end{table*}

\begin{table}[htbp]
\caption{Model Evaluation: Physician Related Entities}
\begin{center}
\begin{tabular}{|c|c|c|c|c|c|c|}
\hline
\textbf{Model}&\multicolumn{3}{|c|}{\textbf{Name}} &\multicolumn{3}{|c|}{\textbf{Address}}\\
\cline{2-7} 
\textbf{Type} & \textbf{\textit{P}}& \textbf{\textit{R}}& \textbf{\textit{F1}}& \textbf{\textit{P}}& \textbf{\textit{R}}& \textbf{\textit{F1}}\\
\hline
Base& 0.56& 0.56& 0.56& 0.45& 0.48& 0.46\\
\hline
Base+Corr& 0.59& 0.53& 0.55& 0.48& 0.46& 0.47\\
\hline
Base+Corr+Post& \textbf{0.73}& \textbf{0.65}& \textbf{0.69}& \textbf{0.69}& \textbf{0.66}& \textbf{0.67}\\
\hline
\end{tabular}
\label{tab:physician}
\end{center}
\end{table}

\begin{table}[htbp]
\caption{Model Evaluation: Exam Related Entities}
\begin{center}
\begin{tabular}{|c|c|c|c|c|c|c|}
\hline
\textbf{Model}&\multicolumn{3}{|c|}{\textbf{Reason For Exam}} &\multicolumn{3}{|c|}{\textbf{Procedures}}\\
\cline{2-7} 
\textbf{Type} & \textbf{\textit{P}}& \textbf{\textit{R}}& \textbf{\textit{F1}}& \textbf{\textit{P}}& \textbf{\textit{R}}& \textbf{\textit{F1}}\\
\hline
Base& \textbf{0.55}& \textbf{0.51}& \textbf{0.53}& 0.48& 0.42& 0.45\\
\hline
Base+Corr& 0.53& 0.49& 0.51& 0.49& 0.45& 0.47\\
\hline
Base+Corr+Post& \textbf{0.55}& \textbf{0.51}& \textbf{0.53}& \textbf{0.52}& \textbf{0.48}& \textbf{0.50}\\
\hline
\end{tabular}
\label{tab:exam}
\end{center}
\end{table}

\section{Results and Discussion}

A comparison of the strategies presented in Section \ref{sec:training}, considering differences in the bounding boxes passed to LayoutLMv3 (word-level, segment-level, or segment+word-level), revealed the worst performance when considering only segment-level bounding boxes. The segment-level model outperformed the word-level bounding box input model only in large entities, namely Address-type and Exam-type entities, but had the worst results of three approaches for the rest of the entities (results not shown). The segment+word-level model improved the performance in longer entities. It performed the best in Address-type and Exam-type long entities, and better than the segment-only model in the rest of the entities. The results shown in this section are for the word-only model, due to its best overall performance, including the best performance in medium and smaller entities out of the three models. For referral management, the patient name, physician name, patient DOB, and patient gender act as important identifiers. 

Experimental results are listed in Table \ref{tab:patient}, Table \ref{tab:physician}, and Table \ref{tab:exam} for the LayoutLMv3 model (Base), the model with corrected labels following the preprocessing of Section \ref{sec:preprocessing} (Base+Corr), and the model with both preprocessing of annotations and postprocessing of predictions (Base+Corr+Post).

The evaluation metric, MUC-5, does not conform to the traditional standard of a confusion matrix, where the denominator of precision and recall produces decreased numbers across the board compared to using the standard f1, precision, and recall calculations. However, these metrics are required to better understand the implications of deploying such a model in a business production setting.

The hybrid model with postprocessing (Base+Corr+Post) resulted in an increase in precision, recall, and F1 scores across all entity types by an average of \textbf{38.5\%}, \textbf{10.06\%}, and \textbf{14.1\%} respectively. This proves that our proposed hybrid approach performs much better than using just the model predictions when we have training data limitations.

For partial matches, most of the F1 scores are above 0.65, with the major identifiers among the top scorers. Analysis of the number of entities predicted per page revealed at most one instance for an entity per page in a majority of cases, likely since the training data has at most one instance of an entity type labeled. However, the curated test dataset has every occurrence of an entity type labeled. Even in the case of identifying any one instance, the results show poor recall compared to precision in all entities. We speculate the single annotation per type per page in the training set generated noise in the model whenever it tried to predict the same entity in any other location other than the one labeled, and the model results may improve with a comprehensively annotated training dataset.

Poor performance in gender is likely due to the single token entity  often missing in the noisy training data and smaller loss contribution from the single-token entity. Comparing Table \ref{tab:patient} and Table \ref{tab:physician}, patient name entity results are better than physician name entity results. The incorrect predictions are due to referrals having multiple unique physician names listed on a single page along with the true referring physician name. Identifying exam procedure and reason for exam presented additional challenges as the entities can be found in a variety of forms, such as descriptive text and/or codes and annotation analysis showed that no single convention was followed.  

Precision was improved with the hybrid model, as the model showed with better partial results that it is good at detecting the location of the entities on a page. We expect completeness and curation of the training dataset annotations will lead to improvements in the recall and F1 scores.

\section{Conclusion}
Information extraction from healthcare referral documents presents multiple challenges due to varied structures and confounding entities. We proposed a hybrid multi-modal model to extract information related to the patient, referring physician, and exam, and evaluated a model trained on incomplete annotations. With MUC-5 F1 scores in the 0.66-0.76 range for most patient and physician entities, we expect this hybrid approach trained on complete and curated annotations can positively impact the referral management process.

\section*{Acknowledgment}
The authors would like to express their sincere gratitude to Yash Soni, Rajat Garg, Rahul Ranjan, and Ankur Goel for the development of the annotation tool. The authors also acknowledge Aakarsh Sethi for his insight in ensuring the study's practical relevance.

\bibliographystyle{IEEEtran} 
\bibliography{ichi_industry}

\end{document}